\documentclass[conference]{IEEEtran}

\usepackage[dvips]{graphicx}
\usepackage{amsmath,amssymb}
\usepackage{algorithm}
\usepackage{pgfplots}
\usepackage{enumerate}
\usepackage{url}
\usepackage[outdir=./]{epstopdf}
\usepackage{flushend} 
\usepackage[most]{tcolorbox}
\usepackage[T1]{fontenc}
\usepackage[utf8]{inputenc}
\usepackage[russian,english]{babel}
\usepackage{algorithmic}
\usepackage[
style=ieee,
isbn=true,
url=true,
defernumbers=false,
sorting=none, 
bibencoding=utf8,
language=auto,    
autolang=other,
]{biblatex}
\usepackage{hyperref}

\begin{document}
\title{Unsupervised cross-lingual matching of product classifications}
\date{}

\author{
\IEEEauthorblockN{Denis Gordeev, Alexey Rey, Dmitry Shagarov}
\IEEEauthorblockA{RANEPA, Russian Presidential Academy of National Economy and Public Administration \\Moscow, Russia \\{rey-ai, gordeev-di, shagarov-dy}@ranepa.ru\\}

}

\maketitle

\begin{abstract}
Unsupervised cross-lingual embeddings mapping has provided a unique tool for completely unsupervised translation even for languages with different scripts. In this work we use this method for the task of unsupervised cross-lingual matching of product classifications. Our work also investigates limitations of unsupervised vector alignment and we also suggest two other techniques for aligning product classifications based on their descriptions: using hierarchical information and translations.
\end{abstract}

\section{Introduction}
Since the works by Bengio et al. \cite{bengio} and Mikolov et al. \cite{mikolov-representations-2013} word embeddings have proven to be an effective and computationally-affordable mechanism to present information about words in a dense vector form and to pass it to neural networks and other classifiers. They are successfully used in many domains and are applied to many state-of-the-art natural language processing models at the first stage of computations \cite{levy-goldberg-2015}.

However, efficient embeddings training requires large amounts of data which might be unavailable for rare languages. Moreover, some tasks like machine translation require a lot of annotated data which are especially scarce for languages other than English. Learning mappings between embeddings from different languages or sources has proven to be a rather efficient method for solving this problem \cite{ruder-survey}.

Thus, Alexis Conneau et al. \cite{muse} have published a programming library called MUSE to map embeddings from two different sources into a single space. They have reached 81.7\% accuracy for English-Spanish and 83.7\% for Spanish English pairs for top-1500 source queries in a completely unsupervised mode. For English-Russian and Russian-English their results are not as high and they achieved accuracy of 51.7\% and 63.7\% respectively. Their FastText embeddings were trained on respective Wikipedia datasets for each corresponding language. 

Artetxe et al. have investigated into limitations of MUSE and show its results to be low for some language pairs, e.g. English-Finnish (0.38\% accuracy). They also present their own solution called Vecmap \cite{vecmap} that outperforms MUSE for this task. For the provided dataset it gets 37.33\% for Spanish-English on average of 10 runs and 37.60\% as the best result (they estimate MUSE result to be 21.23\% on average of 10 runs and 36.20\% at best) and 32.63\% on average for the English-Finnish language pair.

Anders Søgaard et al. \cite{ruder-muse-limitations} also study generative adversarial networks for word embeddings mapping and report that MUSE achieves 00.00\% accuracy for English-Estonian and 00.09\% accuracy for English-Finnish in the unsupervised mode. Moreover, the result for Estonian in the supervised mode is 31.45\% and for Finnish -- 28.01\%. They also report extreme variance for these language pairs (20-30\% difference in accuracy between different random seeds). The authors state that supervision in the form of identically spelled words achieves the same or better results, and thus renders unsupervised methods unnecessary. Another their argument indicates that it is problematic (the performance is close to zero) to map embeddings in the unsupervised way without large corpora, if different algorithms are used for training embeddings(CBOW Word2Vec and Skip-gram Word2Vec) and if embeddings are trained on texts from different domains.

However, it is worth noting that language translation in its pure form is not the only case where algorithms can benefit from using parallel cross-lingual embeddings. In this paper we aim at building an algorithm for unsupervised mapping between coding taxonomies for classifying products. According to the UN \cite{unsd} there are at least 909 (the list seems incomplete - the currently used OKPD2 for Russia is not listed) classifications from 159 countries and most of them except the most prominent ones are unaligned. As examples of such taxonomies we use NIGP-5 and its Russian counterpart OKPD2 \footnote{OKPD2 is a Russian national classification for goods and services introduced in 2014. It has a four-level hierarchy. Categories consist of a code and its description (e.g. 01.11.11.112 - Seeds of winter durum wheat where 01.11.1 code corresponds to Wheat). NIGP-5 is its 2-level US analogue (e.g. 620-80 would be pens and 620 -- office supplies)}.

\section{Related work}

MUSE is based on the work by Conneau et al. \cite{muse}. It consists of two algorithms. The first one which is used only in unsupervised cases is a pair of adversarial neural networks. The first neural network is trained to predict from which distribution $\{X, Y\}$ embeddings come. The second neural networks is trained to modify embeddings $X$ multiplying it by matrix $W$ to prevent the first neural network from making accurate discriminations. Thus, at the end of the training we get a matrix $WX$ which is aligned with matrix $Y$.

The second method is supervised and the aim is to find a linear mapping $W$ between embedding spaces $X$ and $Y$ which can be solved using Orthogonal Procrustes problem:

$$ W^* = argmin_W ||WX - Y||_F = UV^T$$

where $UV^T$ is derived using singular value decomposition SVD$(YX^T) = U \Sigma V^T$
This method is used iteratively with the default number of iterations in MUSE equal to 5.As Søgaard et al. state Procrustes refinement relies on frequent word pairs to serve as reliable anchors.

Conneau et al. also apply cross-domain similarity local scaling to reduce the hubness problem to which cross-lingual embeddings are prone to \cite{dinu}. It uses cosine distance between a source embedding and $k$-target embeddings (the default $k$ in MUSE is 10) instead of the usual cosine distance to generate a dictionary.

$$sim_{source/target} = \dfrac{1}{k}\sum_{i=1}^Kcos(x, nn_i)$$
$$CSLS(x,y) = 2cos(x,y) - sim_{source}(x)  - sim_{target}(y)$$

Vecmap is based on works by Artetxe, Labaka and Agirre. It is close in its idea to the Procrustes refinement, they compute SVD-factorization SVD$(YX^T) = U\Sigma V^T$ and replace $X$ and $Y$ with new matrices $X' = U$ and $Y' = V$. They also propose normalisation and whitening (sphering transformation). After applying whitening new matrices are equal to:
$X' = (X^TX)^{-\tfrac{1}{2}}$ and $Y' = (Y^TY)^{-\tfrac{1}{2}}$

Jawanpuria et al. \cite{jawanpuria} propose a method which is also based on SVD-factorization but in smooth Riemannian manifolds instead of Euclidean space.

\section{Methods and materials}

We approach the problem of matching national product and services classifications as the problem of unsupervised bilingual dictionary induction. In this task vectors corresponding to each category from one classification are aligned with vectors from another classification and classifications itself correspond to languages in the original problem. Several vectors from the first taxonomy may correspond to the same vector from the second taxonomy.

\begin{center}
\begin{table}[!htbp]
\caption{Taxonomy examples}
\begin{tabular}{p{1.5cm}|p{3cm}|p{3cm}}
  Category code & Category description \newline (translated for Russian) & Bid description\\
  \hline
  \hline
 325-25 & Dog and Cat Food & Dog Food: Blue Buffalo Chicken and Brown Rice Food\\
  43.31.10 & \begin{otherlanguage*}{russian}Работы штукатурные\end{otherlanguage*} \newline Plastering Works & Overhaul of the Basement Of The Administration Building\\
\label{table-taxonomies}
\end{tabular}
\end{table}
\end{center}

As examples of such product classifications we consider Russian taxonomy OKPD2 and US NIGP-5.
Both NIGP-5 and OKPD2 are used to classify products and services. However, they differ in the way products are described (two-level vs four-level hierarchy) as well as in the amount of described categories (8700 for NIGP-5 \cite{wiki-nigp} vs 17416 for OKPD2 \cite{wiki-okpd}). It means that two graphs that might describe these product classifications are not isomorphic (contain the same number of graph vertices connected in the same way and may be transformed into one another) by itself.  It does not imply that they may not be made isomorphic by disregarding some vertices (e.g. using some threshold or similarity measure) and then aligned using the methods described above but it complicates their alignment. It should be also noted that some notions from one classification may not exist in the other (e.g. popular in Russia curd snacks and traditional Russian felt footwear 'valenki' do not appear in NIGP-5).

The data for the Russian taxonomy OKPD2 was collected from them Russian state website zakupki.gov.ru \cite{gos-zakupki}, which contains purchases made by state entities.
The data for the US national classification was collected from the US state website data.gov \cite{data-gov}. We have used only marketplace bids by the state of Maryland because they were the only found entries that contained bids descriptions not matching code-descriptions that are required for training Doc2Vec. Extracts from taxonomies can be seen in Table \ref{table-taxonomies}.

Unlike the task of usual cross-lingual matching taxonomy alignment cannot rely on identical strings in category names. Moreover, the task is complicated by the fact that purchases descriptions including categories are collected from absolutely different domains. The task is also affected by the fact that corpora sizes cannot be large (only 70'826 unique entries for NIGP-5 and 1'124'338 -- for OKPD2) because of the lack of data, and thus efficient training of word and document embeddings is hardly possible. Thus, we had to resort to out-of-domain pre-trained embeddings what might convey performance costs. The number of categories is much fewer than the number of words, it carries both advantages and disadvantages (easier to train but our vectors do not contain frequency information which is very important for MUSE) \cite{ruder-muse-limitations}.

Several methods and their combinations were used for mapping taxonomy embeddings.

All studied mapping methods first require word embeddings. We used Doc2Vec \cite{doc2vec} method for getting embeddings describing taxonomies categories. It was trained with library Gensim \cite{gensim}. We have also used pre-trained FastText \cite{fasttext} embeddings provided by the MUSE repository and Google News vectors trained with CBOW Word2Vec \cite{mikolov2013}. 

CBOW Word2Vec is a shallow neural network consisting of two matrices of weights $\underset{V\times d}{\mathrm{W}}$ and $\underset{d\times V}{\mathrm{U}}$ where $V$ is the size of the vocabulary and $d$ is dimensions of the hidden layer used as the word embedding.
The first matrix is used as the embeddings.
The aim of this neural network in its basic variant is to predict the word $w_i$ using its averaged context words
$$v_c = \dfrac{\sum\{w_{i-c},...w_{i-1},...w_{i+1}, w_{i+c}\}}{2c}$$
where $c$ is the window size. The objective function to minimize is

$$ -u_c^Tv_c + \log \sum_{j=1}^{|V|}\exp(u_j^{Tv_c})$$

where $u_c$ is the target word in matrix $U$.

In Doc2vec (PV-DBOW) the model is similar but the aim is to predict the target word using a document's vector (it can be considered just another word).

In FastText words are replaced by n-grams that they consist of, word vectors are computed as averaged n-gram vectors.

We tried several matching techniques:
\begin{enumerate}
\item First we used untranslated texts only:

We trained Doc2Vec on marketplace bids descriptions. After that we used Doc2Vec to get vectors for each category. Using these vectors we trained Vecmap and MUSE in the unsupervised mode with various parameters \{batch sizes - [100,1000], epoch sizes - [100, 1000], number of the Procrustes refinements - [5,1000] \} for MUSE to match vectors for corresponding taxonomies.

\item Translated category descriptions

We translated category descriptions for OKPD2 into English using Google Translate as a proof of concept.
\begin{itemize}
\item MUSE and Vecmap were trained in the unsupervised mode on vectors gained from \begin{itemize}
    \item averaging Word2Vec category descriptions for each taxonomy.
    \item category descriptions for each taxonomy from the English Doc2Vec model trained in the first step
    \end{itemize}
\item Using the averaged Word2Vec vectors received in the previous step we created a dictionary for 10, 30, 50 and 70 \% of matching categories and tried training MUSE and Vecmap in the supervised and semi-supervised (for Vecmap only) mode
\item We found closest strings for category descriptions between different taxonomies. We considered category descriptions from each taxonomy as bags of words, then for a set of words from the first taxonomy we calculated averaged similarity to all category descriptions from the second taxonomy and chose the category from the second taxonomy with the largest similarity. Thus, our method resembles Monge-Elkan similarity \cite[p.~111]{dupe-detect}:
\end{itemize}

$$mapping\{A_i, B\}_{i=1}^{|A|} = max_{j=1}^{|B|}\{sim(A_i,B_j)\}$$

where $$sim = \dfrac{|A_i \cap B_j|}{2} + \\ \dfrac{|B_j \cap A_i|}{2} $$
We used our custom similarity function to fine the function in the cases when the first set of strings is short in comparison with the second set (or opposite).
\begin{itemize}
\item Closest string and Word2vec hierarchical matching (the highest (the most general) category from the source embeddings was matched with the highest category from the target embeddings)
\item We used averaged Word2Vec computed on category descriptions  to find closest strings using cosine distances between the vectors.
\end{itemize}
\item Untranslated texts in the common embedding space.

We used cross-lingual embeddings in a single vector space provided by the authors of MUSE (i.e. vectors for "cat" and its Russian translation \foreignlanguage{russian}{"кот"} are close to each other). Using these embeddings we trained Word2Vec and got averaged vectors for each category using its description.
\item We also used hierarchical information to modify mappings from direct string comparison and averaged Word2vec descriptions. For each category from the first taxonomy we evaluated the closest upper-level category from the second taxonomy. Then we looked for categories that corresponded to the category chosen at the upper level (e.g. if the chosen category code is 64.12 at the next level we look only for categories 64.12.1, 64.12.2).
\end{enumerate}

\begin{center}
\begin{table}
\caption{Illustration of category allignment}
\begin{tabular}{p{1cm}|p{1.5cm}|p{1cm}|p{2cm}|p{1cm}}
  Source \newline Category  \newline Code & Source Category description& Target \newline Category Code & Target Category Description \newline (translated from Russian) & Result\\
  \hline
  \hline
 800-16 & Shoes and Boots: Boots, Rubber &
 43.31.10 & \begin{otherlanguage*}{russian}Сапоги резиновые\end{otherlanguage*} \newline Rubber boots &
 True
 \\
 \hline
  958-78 & Management Services Property Management Services &
  84.11. \newline 19.110 & \begin{otherlanguage*}{russian}Услуги государственного управления имуществом\end{otherlanguage*} \newline State property management services &
  Partially True (state)
  \\
 \hline
 936-70 & Roofing Equipment and Machinery Maintenance and Repair &
 33.12.\newline 23.000 & \begin{otherlanguage*}{russian}Услуги по ремонту и техническому обслуживанию оборудования для металлургии
\end{otherlanguage*} \newline Services in repair and maintenance service of the equipment for metallurgy &
 False

\label{table-annotation}
\end{tabular}
\end{table}
\end{center}

Mappings made by all methods were manually annotated on top-N examples by the corresponding similarity metric (cosine distance for vectors and our string similarity function for strings similarity). If for the first 50 (about top-1\%) examples the accuracy was below 1\% we dropped annotation for this method. The probability for at least one correct example for a random matching method is difficult to estimate (there may be several categories from the second classification that correspond  to the category from the first classification and we do not have the reference alignment) but it may be estimated as ~0.006\%. Otherwise, we annotated top-5\% (231 examples). The annotation included three classes: True, False, Partially true. Partially true examples are usually those that are too specific (fuel management -> nuclear fuel management; rubber shoes -> women rubber shoes; property management services -> state property management) or just not accurate enough according to the assessor. During accuracy estimation "partially true" entries were considered as wrong matches.
\section{Results}

\begin{figure}[!htbp]

	\centering
	\includegraphics[width=0.3\textwidth]{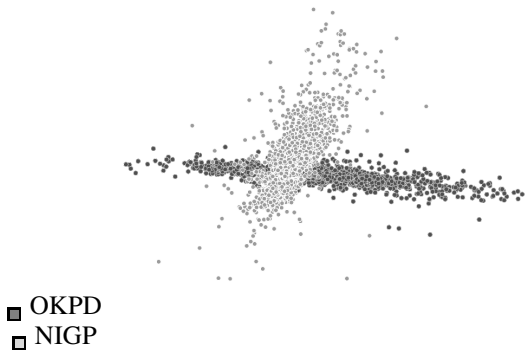}\\
    \raggedright
    \begin{tabular}{c}

    \fcolorbox{black}{gray}{} OKPD \\
    \fcolorbox{black}{gray!30}{} NIGP
	\end{tabular}
	\caption{PCA visualisation of Doc2Vec vectors for OKPD and NIGP-5 taxonomies}
	\label{original-doc2vec}
\end{figure}

\begin{figure}[!htbp]

	\centering
	\includegraphics[width=0.3\textwidth]{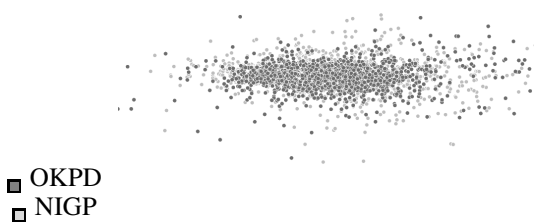}\\
    \raggedright
    \begin{tabular}{c}
    \fcolorbox{black}{gray}{} OKPD \\
    \fcolorbox{black}{gray!30}{} NIGP
	\end{tabular}
	\caption{PCA visualisation of averaged Word2Vec vectors for OKPD and NIGP-5 taxonomies after applying unsupervised MUSE}
	\label{muse}
\end{figure}
\begin{figure}[!htbp]

	\centering
	\includegraphics[width=0.3\textwidth]{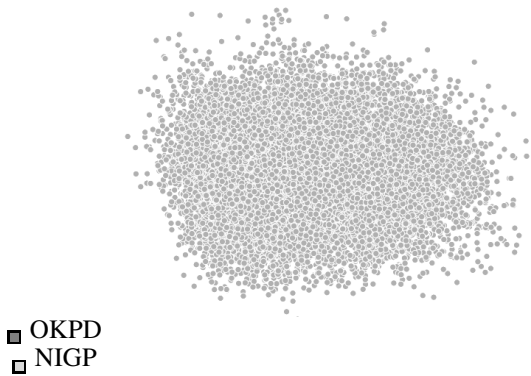}\\
    \raggedright
    \begin{tabular}{c}
    \fcolorbox{black}{gray}{} OKPD \\
    \fcolorbox{black}{gray!30}{} NIGP
	\end{tabular}
	\caption{PCA visualisation of averaged Word2Vec vectors for OKPD and NIGP-5 taxonomies after applying supervised Vecmap with 50\% categories in the dictionary}
	\label{vecmap}
\end{figure}

\begin{figure}[!htbp]

	\centering
	\includegraphics[width=0.3\textwidth]{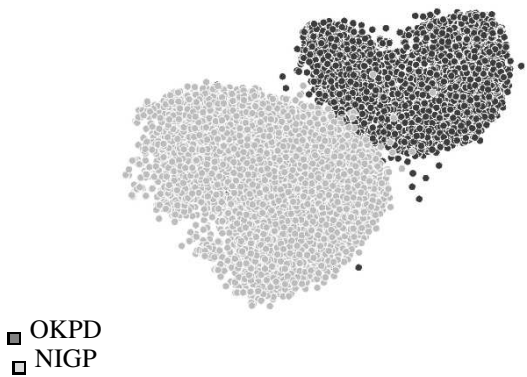}\\
    \raggedright
    \begin{tabular}{c}
    \fcolorbox{black}{gray}{} OKPD \\
    \fcolorbox{black}{gray!30}{} NIGP
	\end{tabular}
	\caption{T-SNE visualisation of original doc2vec embeddings}
	\label{tsne}
\end{figure}

\begin{center}
\begin{table}[!htbp]
\caption{Comparison of matching methods for top-n entries by cosine distance}
\begin{tabular}{p{3.3cm}|p{0.8cm}|p{0.9cm}|p{0.8cm}|p{0.8cm}}
  Method \newline Description & Correct matches & Partially correct matches & Wrong matches &
  Accuracy \newline (\%)\\
  \hline
  \hline
  Translated strings comparison & 126 & 31 & 74 & 55\\
  \hline
  Averaged Word2Vec for translated descriptions & 102 & 53 & 76 & 44\\
  \hline
  Doc2Vec for translated descriptions & 0 & 0 & 50 & 0\\
  \hline
  Unsupervised MUSE with different parameters (none is better) & 0 & 0 & 50 & 0\\
  \hline
  Supervised MUSE with different parameters and reference dictionaries with 1,30,50 and 70\% of the vocabulary (none is better) & 0 & 0 & 50 & 0\\
  \hline
  Vecmap supervised, semi-supervised with various dictionary sizes (10, 30, 50, 70) and unsupervised  & 0 & 0 & 50 & 0\\
    \hline
  averaged Word2Vec using cross-lingual embeddings in single space  & 45 & 19 & 167 & 19.5\\
    \hline
    Hierarchical string comparison & 48 & 40 & 143 & 20.8\\
    \hline
    Hierarchical averaged Word2vec & 94 & 43 & 94 & 40.7\\
    \hline
    Hierarchical averaged Word2Vec using cross-lingual embeddings in single space  & 108 & 7 & 116 & 47.5\\
\label{table-accuracies}
\end{tabular}
\end{table}
\end{center}

In the Table 3 we can see annotation results for top-n best matches according to cosine distances between vectors and the string similarity score for the translation method.

Figure \ref{tsne} shows that original embeddings (Fig. \ref{original-doc2vec}) can be successfully clustered using t-SNE. Vecmap and MUSE manage to align both spaces (Fig. \ref{muse} and Fig. \ref{vecmap}) successfully so that they are not linearly separable. However, as can be seen from Table \ref{table-accuracies} unsupervised matching techniques fail to properly align taxonomy embeddings. MUSE and Vecmap failed to achieve accuracy above 0\%. Word2Vec and string matching demonstrate better results at the level of alignments for low-resource languages reported by Conneau and Søgaard. Results from averaged Word2vec after alignment are worse than those of translated string comparison which may be attributed to pre-trained embeddings being from a different domain. Moreover, averaging tends to make to broad assumptions (thematic in nature) which is unsuitable for the current task. Doc2vec unsurprisingly gets worse results because of the lack of training data. Unfortunately, for low-resource languages string matching is impossible without a working translation solution which unsupervised cross-lingual dictionary alignment strives to solve.

The surprising fact that Vecmap and Muse cannot align data even in supervised and semi-supervised modes with dictionaries created after Word2Vec or string-alignment matching. It can be partially attributed to the insufficient accuracy of the provided dictionaries or a very low amount of categories (in comparison to words). However, it is possible that both methods latently and mainly depend on word frequencies and other similar distributive information provided by word embeddings. Also supervised mapping adjustment turned out to be strong for our dataset.

Using pre-aligned cross-lingual embeddings might appear to be helpful and is useful for rare languages which lack efficient translation engines. Thus, the procedure would be: first, to train word-embeddings using some corpora from a common domain (e.g. Wikipedia), then align them using MUSE or Vecmap. After that, those embeddings may be used to map category descriptions. It should be also noted that mappings annotated as wrong were not completely incorrect and were usually on topic (e.g. acids -> oils; engine maintenance -> auto body repair; sewage treatment equipment -> sewage treatment services). Thus, some other procedure rather than Word2vec averaging might demonstrate better results for this task.

According to our results, it seems unlikely that unsupervised matching techniques might result in sufficiently good dictionary alignment and machine translation for rare languages (e.g. those that do not have rich corpora like  Wikipedia and currently there are only 61 languages with the number of Wikipedia articles exceeding 100'000).

As with the work by Søgaard string matching techniques perform better than their unsupervised counterparts.

Using domain knowledge and hierarchical information turned out to be helpful, especially in the case of pre-aligned vectors. However, hierarchical matching techniques show worse results for averaged Word2Vec and string similarity evaluated on translated category descriptions. For Word2Vec the results are insignificant (the p-value between hierarchical and non-hierarchical version for Fisher test is only 0.5). However, domain information may be even harmful for string matching (the p-value is less than 0.001). It may be explained by the fact that upper-level categories have two broad names and thus it leads to mistakes at lower hierarchy grades. Using hierarchy information is extremely helpful for pre-aligned vectors in a common space(p-value < 0.001). It removes the problem of being too general and increases accuracy. So for structured datasets like Wikipedia it may be helpful to include not only information about word distributions but also meta-information like connections between articles and their hierarchy.

\section{Conclusion}
In this work we have created an unsupervised way to match unaligned national classification systems. We demonstrate that using translation information from a pre-trained translation engine or using embeddings pre-aligned in a common space may help in solving this task. However, it seems unlikely that it is possible to directly align categories vectors for national taxonomies because their domains are too different. Moreover, it turns out that even supervised matching techniques relying on partially matched dictionaries fail at this task. It may be attributed to the low number of categories.

It seems unlikely that it is possible to unsupervisedly match national taxonomies for rare languages which lack any translation engines because general adversarial networks and analytical methods fail to properly align the studied manifolds. Moreover, we support issues raised by Søgaard et al. \cite{ruder-muse-limitations} and demonstrate that both MUSE and Vecmap do not achieve acceptable results both in the supervised and unsupervised mode for tiny datasets from different domains.
The results also hint on the idea that unsupervised dictionary alignment results may be so successful because of the parallel nature of Wikipedia (some articles may be direct translations of English ones). However, it requires further investigation.
We also demonstrate that using structural and hierarchical dataset information may considerably improve matching results, what is applicable to many Internet-based datasets like Wikipedia.

\printbibliography

\end{document}